\pdfoutput=1

\documentclass[10pt, a4paper]{article}

\usepackage[final]{lrec-coling2024} 

\usepackage{times}
\usepackage{latexsym}
\usepackage{amsmath}
\usepackage{amssymb}

\usepackage[T1]{fontenc}

\usepackage[utf8]{inputenc}

\usepackage{microtype}

\usepackage{inconsolata}

\usepackage{algpseudocode}
\usepackage{algorithm}
\usepackage{hyperref}

\title{Neural Multimodal Topic Modeling: A Comprehensive Evaluation}

\name{Felipe González-Pizarro, Giuseppe Carenini}
\address{ Department of Computer Science, University of British Columbia \\Vancouver, BC, Canada \\
         \{felipegp, carenini\}@cs.ubc.ca}

\usepackage{multirow}
\usepackage{graphics}
\usepackage{graphicx} 
\usepackage{verbatim}
\usepackage{subcaption}

\usepackage{arydshln}
\usepackage{import}

\usepackage{color,colortbl,soulutf8}
\usepackage{float}

\definecolor{Yellow}{RGB}{252, 243, 207}
\definecolor{Red}{RGB}{255,204,204}
\definecolor{white}{RGB}{255,255,255}

\usepackage{booktabs}

\usepackage{ifthen}
\newif\ifshort
\shorttrue   
\ifshort
	\newcommand{\isShort}{true}
\else
	\newcommand{\isShort}{false}
\fi
\newcommand{\shortVer}[1]{\ifthenelse{\equal{\isShort}{true}}{{#1}}{}}
\newcommand{\longVer}[1]{\ifthenelse{\equal{\isShort}{false}}{{#1}}{}}

\ifshort
\usepackage[]{flushend}
\else
\fi

\newif\ifcomment
\commentfalse
\ifcomment
\newcommand{\fg}[1]{{\bf\textcolor{purple}{FG: #1}}}
\newcommand{\new}[1]{{\textcolor{blue}{#1}}}
\newcommand{\rl}[1]{{\bf\textcolor{blue}{RL: #1}}}
\newcommand{\gc}[1]{{\bf\textcolor{violet}{GC: #1}}}

\else
\newcommand{\gc}[1]{}
\newcommand{\fg}[1]{}
\newcommand{\rl}[1]{}
\newcommand{\new}[1]{#1}

\fi
\newcommand{\descr}[1]{\smallskip\noindent\textbf{#1}}

\usepackage{comment}
\usepackage{tabularx,booktabs}
\newcolumntype{b}{X}
\newcolumntype{s}{>{\hsize=.03\hsize}X}

\abstract{\new{Neural topic models can successfully find coherent and diverse topics in textual data. However, they are limited in dealing with multimodal datasets (e.g., images and text). This paper presents the first systematic and comprehensive evaluation of multimodal topic modeling of documents containing both text and images. In the process, we propose two novel topic modeling solutions and two novel evaluation metrics. Overall, our evaluation on an unprecedented rich and diverse collection of datasets indicates that both of our models generate coherent and diverse topics. Nevertheless, the extent to which one method outperforms the other depends on the metrics and dataset combinations, which suggests further exploration of hybrid solutions in the future. Notably, our succinct human evaluation aligns with the outcomes determined by our proposed metrics. This alignment not only reinforces the credibility of our metrics but also highlights the potential for their application in guiding future multimodal topic modeling endeavors. }
 \\ \newline }

\begin{document}
\maketitleabstract

\section{Introduction}

%


The vast amount of text that is 
constantly generated has led to the development of several algorithms designed to 
interpret and  summarize large sets of documents ~\cite{peter2015topicks}. A well-known automatic mechanism is topic modeling, a robust approach for extracting core themes 
from large text corpora. In practice, when 
topic modeling 
is applied to a corpus 
(e.g., news articles),
the results 
will include a set  of topics. Usually, each topic is represented by a list of related terms 
(e.g., \textit{tropical, storm, hurricane, cyclone, weather, rain})~\cite{survey_topic_modeling}. Domain experts (e.g., journalists, physicians, and marketers) can use topic modeling 
to analyze 
large document collections without reading every document
~\cite{boyd2017applications,ge2019octvis}.





Since most topic modeling algorithms have been designed specifically to process textual data, their performance is limited in corpora containing information in other modalities (e.g., images and videos). Some work has attempted to address this limitation by expanding well-known probabilistic~\cite{10.1145/2505515.2505652, 9696359} or neural topic models~\cite{zosa2022multilingual} to multimodal settings, especially considering images in the documents. Yet, it remains unclear which method works better for which dataset, and what evaluation metrics are more appropriate in this new multimodal scenario. This paper addresses this gap by conducting the first systematic and comprehensive evaluation of multimodal topic modeling applied to documents containing both text and images. In this process, we make several contributions with respect to neural topic modeling algorithms, evaluation metrics, and datasets.

\descr{Neural topic modeling algorithms:} We 
have developed two novel neural multimodal topic modeling algorithms 
by adapting  
SOTA  solutions\footnote{Our code is available at : \url{https://github.com/gonzalezf/multimodal_neural_topic_modeling/}}. First, we extend ZeroShotTM~\cite{bianchi-etal-2021-cross}, a neural  topic model that only uses pre-trained textual embeddings,  to the multimodal \texttt{Multimodal-ZeroShotTM} 
, which additionally embeds images. In particular, both the Bag-Of-Words (BOW) and the Image Features associated with each document are reconstructed during decoding. 
Secondly, we present 
\texttt{Multimodal-Contrast}, derived from M3L-Contrast~\cite{zosa2022multilingual}, a recent multimodal multilingual neural topic model that uses Contrastive Learning to map texts from multiple languages and images into a shared topic space. Our \texttt{Multimodal-Contrast} simply omits  the encoder and inference networks associated with a second language. 
 
\descr{Metrics:} 
The quality of a given set of topics  can be automatically assessed  primarily based on their coherence and diversity. For topic modeling methods 
that only process textual data, coherence metrics such as NPMI~\cite{lau2014machine}, $C_{v}$~\cite{10.1145/2684822.2685324}, and WE~\cite{10.1145/2911451.2914729} evaluate the semantic relatedness of the topic keywords; while diversity 
metrics like TD~\cite{dieng2020topic} and I-RBO~\cite{bianchi-etal-2021-pre} measure the lexical overlap between the descriptors of different topics. 
However, in a multimodal setting, 
each topic is represented not only by a set of keywords, but also by a set of images. Yet, there are 
no automatic metrics 
to assess the coherence and segregation of the images 
representing a topic. In this paper, to fill this gap, we propose two new metrics, namely Image Embedding-based Coherence (\texttt{IEC}) and Image Embedding-based Pairwise Similarity (\texttt{IEPS}), which appear to align with human judgment in our preliminary user study. 

\descr{Datasets:} While previous work on multimodal topic modeling has been tested only on a few rather homogeneous datasets, we are the first to propose and leverage  six diverse datasets that vary substantially in terms of the document size (ranging from 6  to 2,425 words per document on average), the source of the documents (e.g., Flickr, Twitter, Wikipedia), the underlying task/domain (e.g., Object recognition, Visual Storytelling) as well as in the way the gold-standard data was collected (e.g., crowd-sourcing, automatic classification).

Armed with a comprehensive set of metrics and a diverse collection of datasets, we perform the first systematic 
evaluation of multimodal neural topic modeling methods, comparing our 
novel proposals, 
\texttt{Multimodal-ZeroShotTM} and  \texttt{Multimodal-Contrast},  among themselves and against only textual SOTA topic modeling methods.



\section{Related Work}
\label{sec:related_work}
With the recent developments of deep neural networks, several \textit{Neural Topic Models} (NTMs) have  been proposed. 
For instance, \citet{srivastava2017autoencoding} proposed 
\textit{Product-of-Experts LDA} (ProdLDA),  a topic modeling algorithm that, like the original (LDA)~\cite{blei2003latent}, still uses a BOW representation of documents, but leverages it in a more sophisticated way by combining  
a variational autoencoder (VAE)~\cite{blei2017variational}  with a Product of Experts (PoE) approach ~\cite{hinton2002training}. 
As a result, ProdLDA not only  consistently identifies more coherent and diverse 
topics than LDA~\cite{srivastava2017autoencoding,sridhar-etal-2022-heterogeneous}, 
but it 
also process  data more efficiently ~\cite{srivastava2017autoencoding}.
Despite progress, both LDA and ProdLDA are still limited to BOW document representations. By ignoring critical syntactic and semantic relationships among words, they sometimes fail to identify high-quality topics (see ~\cite{bianchi-etal-2021-pre} and ~\cite{burkhardt2019decoupling}). 
In order to incorporate semantic relationships into topic models, \citet{dieng2020topic} proposed \textit{Embedding Topic Models} (ETM), a generative probabilistic model that relies on static word embeddings~\cite{mikolov2013distributed} to identify interpretable topics. 


Nevertheless, a remaining key shortcoming of ETM is that by relying on static embeddings, it does not consider contextual relations among words. This limitation was recently addressed  by 
\citet{bianchi-etal-2021-pre} and \citet{bianchi-etal-2021-cross}, which proposed \textit{Contextualized Topic Models} (CTM). 
CTM is a family of neural topic models based on a variational autoencoder (VAE) (i.e., CombinedTM~\cite{bianchi-etal-2021-pre}, ZeroShotTM~\cite{bianchi-etal-2021-cross}), that  relies instead on contextual embeddings (e.g., SBERT~\cite{reimers-gurevych-2019-sentence}), obtaining higher quality topics than all previous approaches. One of the two neural topic models we propose in this paper, \texttt{Multimodal-ZeroShotTM}, is based on this recent work. 

In contrast to the aforementioned neural topic models, BERTopic~\cite{grootendorst2022bertopic}, takes a distinct approach by employing a clustering methodology. Unlike the neural topic models that infer a mixture of topics within documents, BERTopic assumes each document correlates with a single topic. This model incorporates heuristic techniques to manage this limitation, yet the effectiveness of these strategies remains an area for further exploration. Our investigation concentrates on neural topic models, which inherently consider the presence of multiple topics in documents, providing a broader understanding of the data's thematic structure.

Only a few topic modeling algorithms have been proposed to process more than textual data. A notable exception is  M3L-Contrast~\cite{zosa2022multilingual}, a neural topic model that maps texts from multiple languages and images into a shared topic space by using pre-trained  image (CLIP~\cite{radford2021learning}), and text embeddings (SBERT~\cite{reimers2019sentence}) to abstract 
the complexities between 
different languages and modalities. The second neural topic model we propose here, 
\texttt{Multimodal-Contrast}, is based on this recent work.

Given the recent success of decoder-only GPT-like systems \cite{openai2023gpt4} in so many NLP tasks, 
it may seem surprising that they have not been applied yet to  the topic modeling task. However, the plain reason is that they are still severely limited in their input size, currently in the 10,000s of tokens~\cite{bubeck2023sparks}, and therefore cannot process the large corpora for which topic modeling is actually needed.
Very recent work (e.g., \cite{yu2023megabyte}) may inspire ideas on how to overcome this limitation in the future. Tellingly, the evaluation framework and baselines presented in this paper will be critical in assessing these new solutions.



\section{Our New Multimodal 
Algorithms}
\label{sec:proposal}




\subsection{Multimodal-ZeroShotTM}

\begin{figure}[!t]
\centering
\includegraphics[width=0.53\columnwidth]{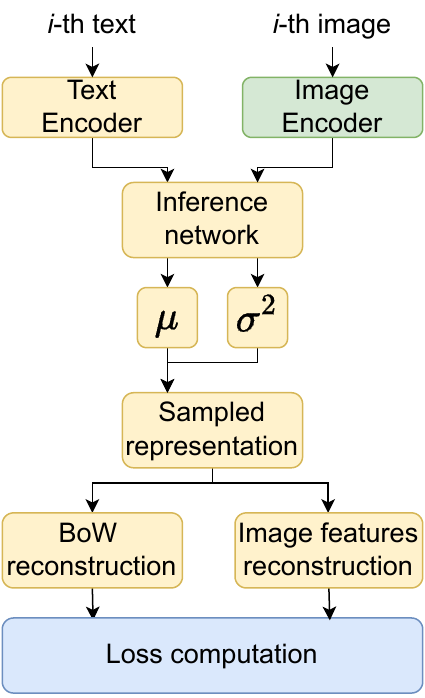}
\caption{High-level schema of the architecture for \texttt{Multimodal-ZeroShotTM}. The loss function is detailed in Equation 
\ref{eq:multimodalctm}.}
\label{fig:MultimodalZeroShotTM}
\end{figure}

We propose \texttt{Multimodal-ZeroShotTM}, a novel multimodal topic modeling algorithm based on ZeroShotTM~\cite{bianchi-etal-2021-cross}. 
Figure \ref{fig:MultimodalZeroShotTM} shows the architecture of our model. Given a document with a textual and visual component (e.g., an image and its caption), we encode each element using a modality-specific encoder (e.g., by using CLIP image and text encoders). Then, we concatenate those embeddings and pass them to an inference network as input. 
After that, the model samples a latent representation from a Gaussian distribution parameterized by $\mu$ and $\sigma^2$, as is done in related works~\cite{bianchi-etal-2021-cross,zosa2022multilingual,srivastava2017autoencoding}. The crucial difference between our model and ZeroShotTM 
is that our decoder network reconstructs both the BoW and the Image Features associated with each document (bottom right of Figure \ref{fig:MultimodalZeroShotTM}). Advantageously, by embedding images and reconstructing their 
features, the model can capture complementary information not present in the textual 
parts of the document, thus plausibly identifying topics that are more representative of the multimodal corpus. 
Our per document loss function $\mathcal{L}$ now includes three components: 
\begin{equation}
\begin{split}
\mathcal{L}& =
\mathbb{E}_q\left[\mathbf{w}^{\top} \log \left(\operatorname{softmax}\left(\beta \theta\right)\right)\right]- \\
&  
\mathbb{K} \mathbb{L}\left(Q\left(\theta \mid \mathbf{x}\right) \| P\left(\theta\right)\right)+ 
 \lambda 
(1-cos(\mathbf{x}_{img}, \gamma \theta)) 
\end{split}
\label{eq:multimodalctm}
\end{equation}

where the first term measures the loss associated with the BoW vector reconstruction of 
the document 
(see \cite{srivastava2017autoencoding} for more details). 
%
The second term corresponds to the sum of the \new{Kullback-Leibler divergence} (KL) loss between the posterior 
and  prior distributions for 
the document embedding $\mathbf{x}$ (i.e., the concatenation of the text and image embeddings). 
The mean $\mu$ and variance $\sigma^2$ of the posterior distributions are estimated in each inference network,  
and $\theta$ is the sampled topic distribution per document embedding. 
Finally, the third term corresponds to the loss associated with the reconstruction of Image Features, as 
the cosine embedding loss\footnote{Using MSE as this 
loss delivers similar performance.} between the  estimated image features $\gamma \theta$ and the actual value $x_{img}$. This loss 
measures the similarity between the 
two vectors 
and is often used for learning nonlinear embeddings.  $\lambda$ is a parameter 
to explore the trade-off between textual and image losses but has been kept equal to 1 in the main experiments.


\subsection{Multimodal-Contrast}
\label{sec:proposal_m3l_contrast}
\begin{figure}[!t]
\centering
\includegraphics[width=0.74\columnwidth]{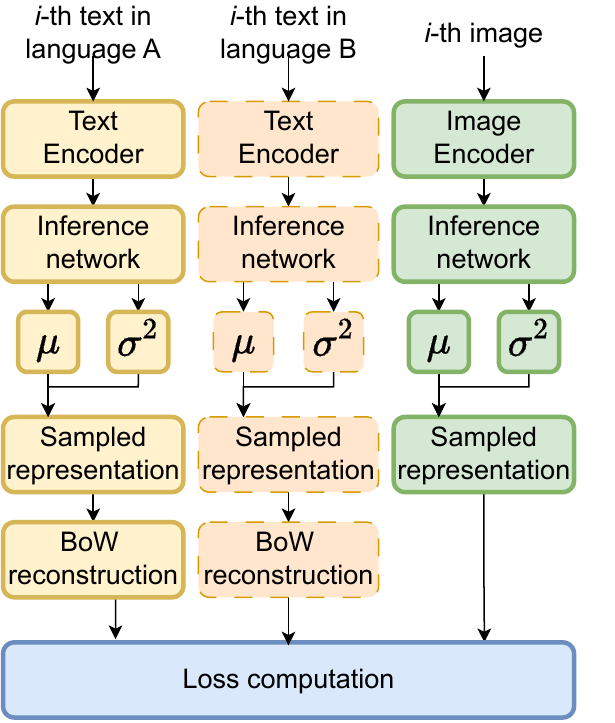}

\caption{M3L-Contrast topic model architecture. It includes language-specific and modality-specific encoders and inference networks. We highlight with a dashed line (- - -) the components that we removed during our adaptation. }
\label{fig:M3L_original_architecture}
\end{figure}

We propose \texttt{Multimodal-Contrast}, an adaptation specifically derived from M3L-Contrast~\cite{zosa2022multilingual}. While M3L-Contrast is a neural topic modeling technique tailored for analyzing datasets that are both multilingual and multimodal, \texttt{Multimodal-Contrast} shifts the focus to solely multimodal data. In  M3L-Contrast,  
each document must contain an image and textual content in two languages (e.g., English and German), with the model's architecture including three encoders and inference networks (see Figure \ref{fig:M3L_original_architecture}), each one processing either  text in one of the two languages or an image. In our adaptation, we essentially removed from the architecture the encoder and inference network for one of the two languages 
(dashed  
in Figure \ref{fig:M3L_original_architecture}, i.e., the components for language B).  

The main difference between \texttt{Multimodal-Contrast} and  \texttt{Multimodal-ZeroShotTM} is 
the third component of the loss function  $\mathcal{L}$. In particular, while \texttt{Multimodal-ZeroShotTM} considers the loss associated with the reconstruction of Image Features, \texttt{Multimodal-Contrast} uses the  InfoNCE Contrastive Learning loss~\cite{oord2018representation} to align topic distributions sampled from the different modalities of the document (i.e., its textual and visual parts). This Contrastive Learning loss maps similar instances close to each other and keeps non-related instances apart. Overall, 
$\mathcal{L}$ 
combines three components: 
%
\begin{equation}
\resizebox{0.90\columnwidth}{!}{
$
\begin{split}
\mathcal{L}& =
\mathbb{E}_q\left[\mathbf{w}^{\top} \log \left(\operatorname{softmax}\left(\beta \theta_{txt}\right)\right)\right]- \\
& 
\sum_{\substack{l=1}}^k  \mathbb{K} \mathbb{L}\left(Q\left(\theta^l \mid \mathbf{x}^l\right) \| P\left(\theta^l\right)\right)- \\
& \omega 
\sum_{\substack{a, b=1 \\
a \neq b}}^k \log \frac{\exp \left(\left(\theta^a \cdot \theta^b\right) / \tau\right)}{\sum_{j=1}^N \sum_{c,d=1}^k \exp \left(\left(\theta^c \cdot \theta_j^d\right) / \tau\right)}
\end{split}
$
}
\label{eq:m3l_loss}
\end{equation}

The first term corresponds to the standard BoW reconstruction loss given the textual component of the document. 
The second term corresponds to the 
KL loss between posterior and prior distributions for every component $k$ (i.e., image, text) of a document. So, it is similar but not the same as in \texttt{Multimodal-ZeroShotTM}. 
Here, the mean $\mu$ and variance $\sigma^2$ of the posterior distributions are estimated in each inference network;  $\mathbf{x}$ can be a textual or visual embedding, and $\theta$ is a sampled topic distribution given a textual or visual component of a document. 
Finally, the third term is the InfoNCE loss, where ($\theta^{a} \cdot \theta^{b}$) are positive pairs and ($\theta^{c} \cdot \theta^{d}$) are negative pairs. Positive terms are aligned components of a document (e.g., an image and its caption). $N$ is the batch size, $\tau$ is the temperature, and $\omega$ is a parameter (like $\lambda$ for \texttt{Multimodal-ZeroShotTM}) 
to explore the trade-off between 
contrastive and other losses, but has been kept equal to 100 in the main experiments as \cite{zosa2022multilingual}.



%



\section{Automatic Metrics}

The quality of topic models is commonly assessed based on their coherence and diversity. 
Automatic coherence metrics  identify the degree of lexical and semantic relatedness between the terms that describe each topic, while automatic diversity metrics measure the lexical overlap between the terms of different topics. In our evaluation, we apply standard metrics to assess the quality of the textual descriptors 
of the topics, namely NPMI~\cite{lau2014machine}, $C_{v}$~\cite{10.1145/2684822.2685324}, and WE~\cite{10.1145/2911451.2914729} for coherence and TD~\cite{dieng2020topic} and I-RBO~\cite{bianchi-etal-2021-pre} for diversity\footnote{We use the implementations of these metrics provided in the OCTIS library~\cite{terragni2021octis}.}. 

However, in multimodal topic models, each topic is represented not only by a set of keywords but also by a set of  images. Since there are currently no automatic metrics for evaluating the coherence and diversity of images representing topics, we propose  two new metrics, namely \texttt{Image Embedding-based Coherence (IEC)} and \texttt{Image Embedding-based Pairwise Similarity (IEPS)}, 
to fill this gap. 

\begin{table*}[!ht]
\centering
\small
\resizebox{0.99\textwidth}{!}{%
\begin{tabular}{p{0.15\textwidth}p{0.15\textwidth}
p{0.15\textwidth}p{0.15\textwidth}p{0.15\textwidth}p{0.15\textwidth}}
\toprule
\multicolumn{1}{c}{MS COCO} & \multicolumn{1}{c}{VIST} & \multicolumn{1}{c}{T4SA} & \multicolumn{1}{c}{MMHS150K} & \multicolumn{1}{c}{HC-4chan} & \multicolumn{1}{c}{MEWA} \\
\midrule
                        \includegraphics[width=0.16\textwidth]{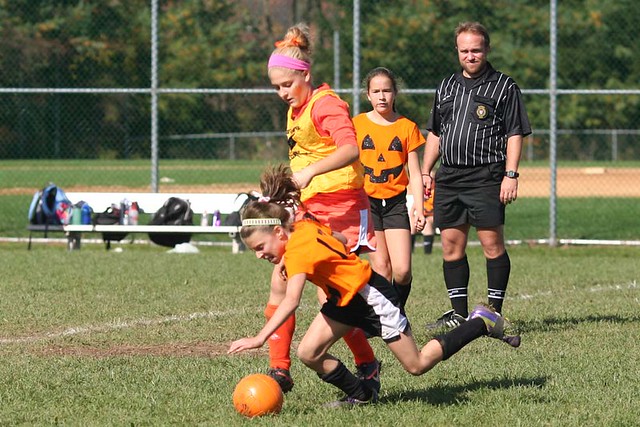}
                        &
                        \includegraphics[width=0.16\textwidth]{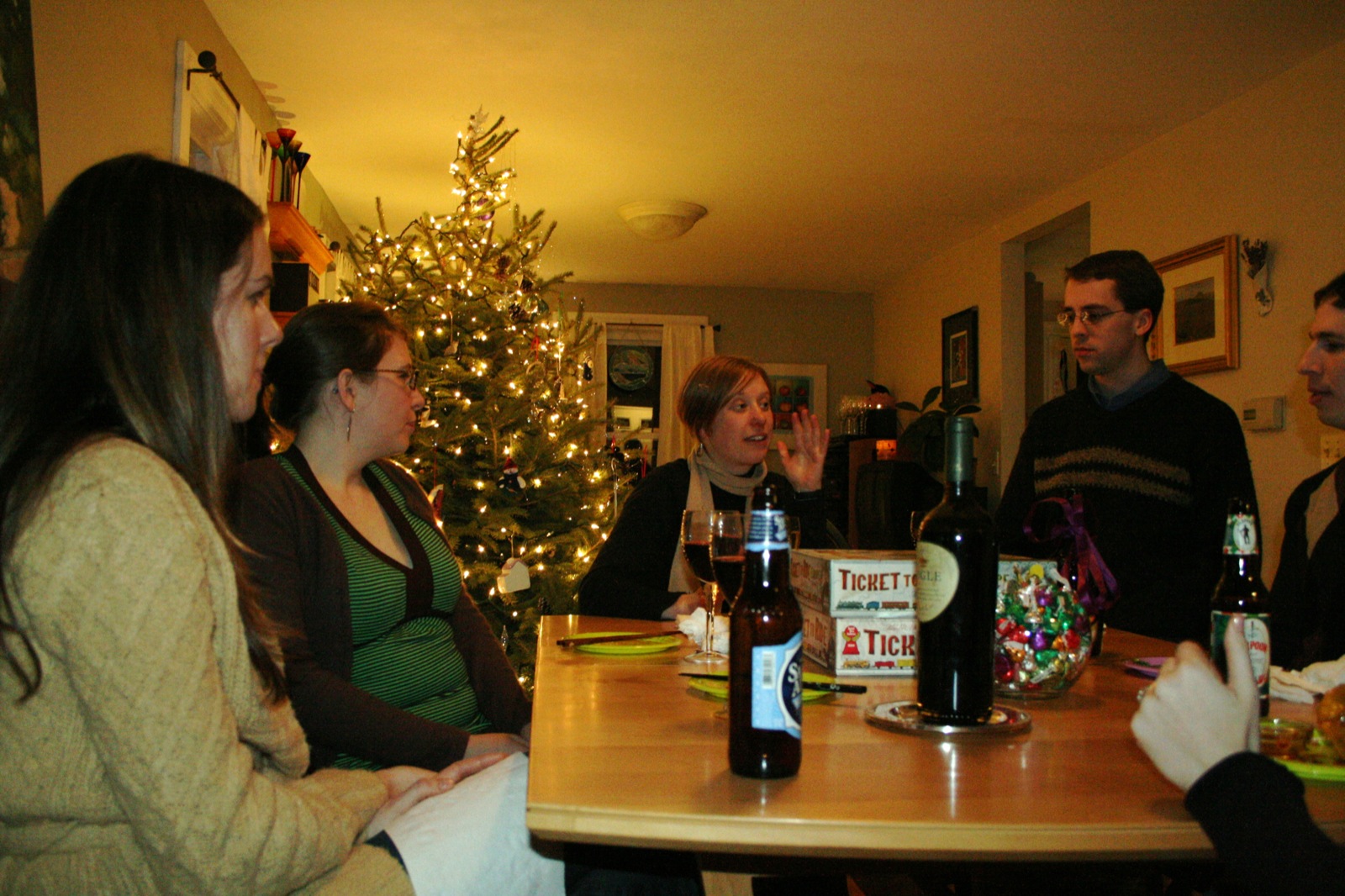} &  
                    \includegraphics[width=0.16\textwidth]{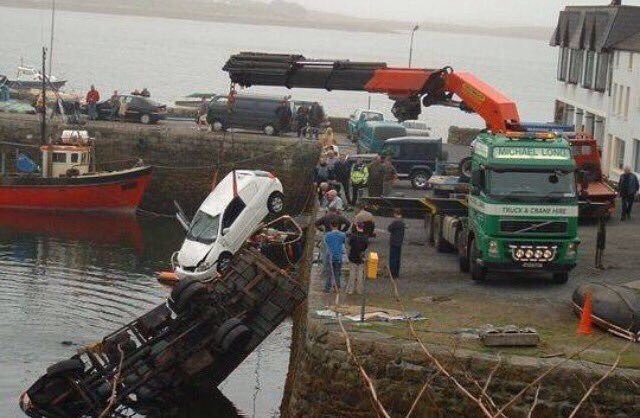} &  
              \includegraphics[width=0.16\textwidth]{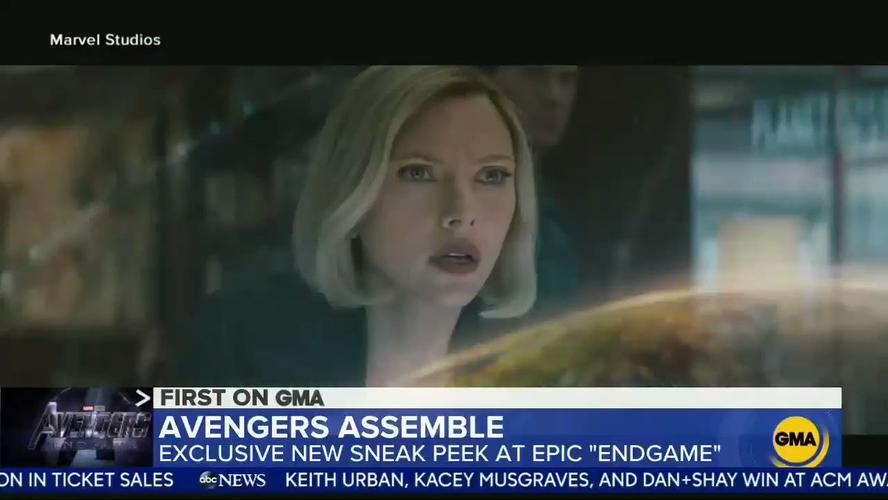}  &     
                    \includegraphics[width=0.1\textwidth, height=0.1\textwidth]{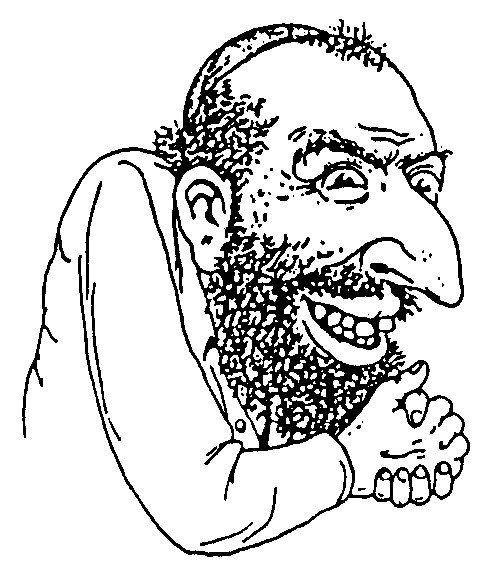}
                &            \includegraphics[width=0.16\textwidth]{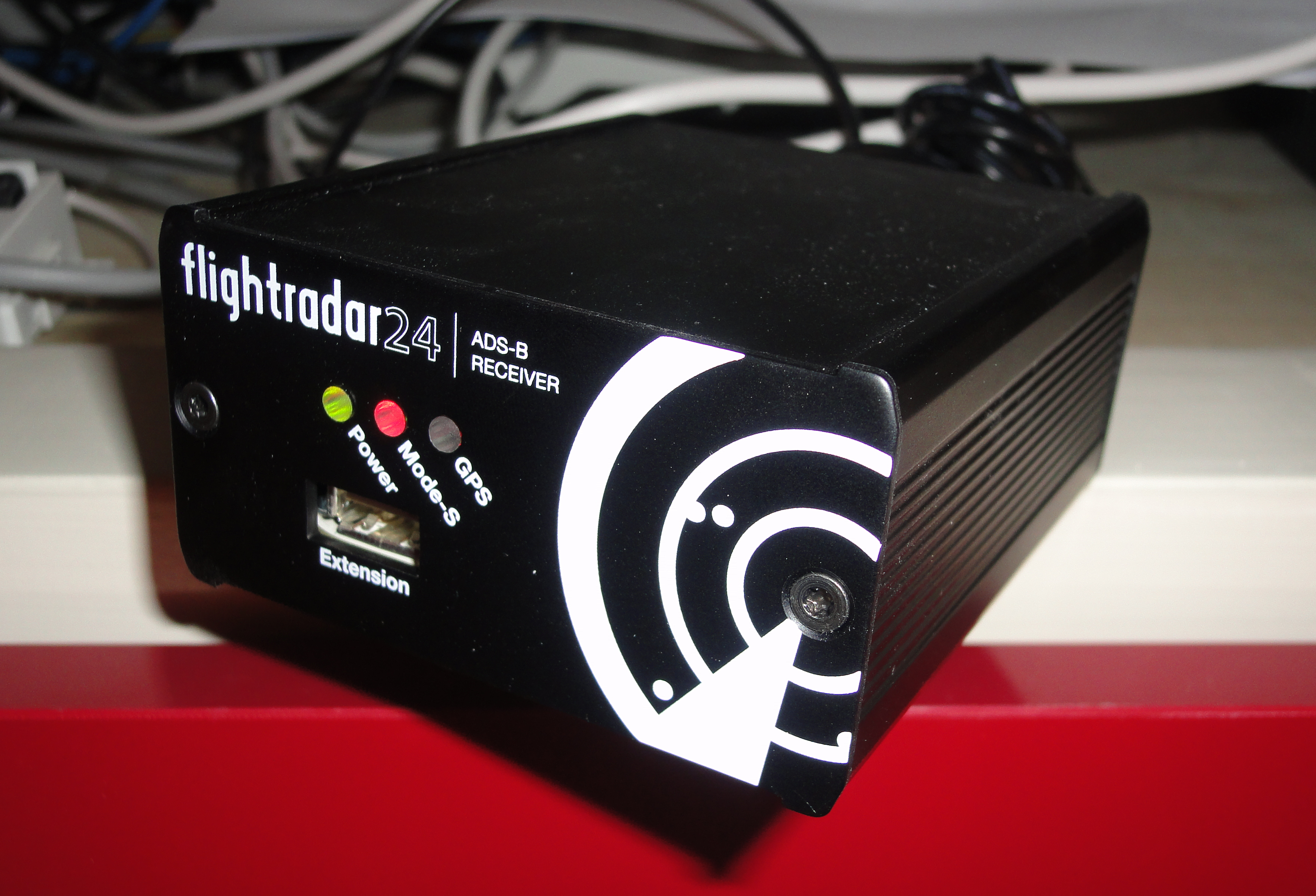}                       \\

          there are some children playing a soccer game
                         &        
                         
                        Having a good time bonding and talking.	&  
                        RT @AmBlujay: This is why l won't interfere in people's relationships   &
                                            carol really said f*ck yall I’m a d*ke and I’m here to save the universe &
                         Yes goyim! Don't fight it! It's inevitable                                                  
                                  &         Flightradar24 is a Swedish internet-based service that...               \\
\bottomrule
\end{tabular}
}
\caption{Dataset document examples. To save space, we only show the first terms of the document from  MEWA. \new{\textbf{Disclaimer:} HC-4chan and MMHS150K contains hateful textual and graphic elements.}}
\label{tab:datasets_example}
\end{table*}

\descr{Image Embedding-based Coherence (IEC)} 
is based on WE~\cite{10.1145/2911451.2914729}, a metric that has been validated and widely used~\cite{bianchi-etal-2021-pre, li-etal-2023-diversity}. While the WE metric was designed to gauge the semantic relatedness between word embeddings, our IEC
evaluates the semantic relatedness between images representing a topic. 
Formally, let $T$ be the set of topics, and $W_t$ be the set of top-\new{\textit{N}} images in topic $t \in T$, 
The average pairwise image similarity for topic $t$ is:
\begin{equation}
\text{sim}(W_t) = \frac{1}{\binom{|W_t|}{2}} \sum_{i=1}^{|W_t|} \sum_{j=i+1}^{|W_t|} \text{cosine}(w_i,w_j)
\end{equation}
where 
$\text{cosine}(w_i,w_j)$ computes relatedness between any two images $w_i$ and $w_j$ as the cosine similarity between their corresponding embeddings.
Finally, our new metric, IEC, is simply the average of the topic-level similarity scores across all topics:
\begin{equation}
\text{IEC} = \frac{1}{|T|} \sum_{t \in T} \text{sim}(W_t)
\end{equation}
\new{IEC} ranges between [0,1], where a higher value suggests more coherent topics.
\begin{table*}[!ht]
\centering
\resizebox{\textwidth}{!}{%
\begin{tabular}{@{}lrrrcp{0.4\textwidth}c@{}}
\toprule
Dataset &   \# Docs  & Vocab size & Avg. Len & Domain &  Data collection & Source\\
\midrule
 
 MS COCO &   30,000 &   2,000 &10.46 & Object recognition & Tag-matching \& crowdsourcing & Flickr\\
 VIST &  30,000 &   2,000 &9.99 & Visual storytelling & Crowdsourcing & Flickr   \\
 T4SA &   30,000 &    2,000  & 6.99 & Sentiment analysis  & Automatic sentiment classification& Twitter  \\
 MMHS150K &   30,000 &   2,000  & 6.03 &  Hateful content detection& Keywords-matching \& crowdsourcing & Twitter   \\
 HC-4chan &   17,866 &    1,993 & 17.43 & Hateful content detection & Human annotations \& automatic detection & 4chan  \\
 MEWA &   18,592  &  2,000  &  2,424.83&  Information retrieval & Crowdsourcing & Wikipedia    \\ 
 \bottomrule
\end{tabular}
}
\caption{Properties of the datasets used. 
Vocab size is restricted to 2000 words to speed computation}
\label{tab:datasets_stats}
\end{table*}

\descr{Image Embedding-based Pairwise Similarity (IEPS)} 
measures the diversity of a topic model by computing the similarity between all the topics and then considering lower scores as a sign of higher diversity.  
Formally, by adapting the Word Embedding-Based Pairwise Similarity metric (WEPS)~\cite{terragni2021word}  (\new{a validated metric} originally designed for \new{assessing the similarity between word} 
embeddings), we first define the similarity between  the top-\textit{N} images that describe any two topics $t_i$ and $t_j$ as follows: 
\begin{equation}
\text{IEPS}_{\text{pair}}{(t_i,t_j)} = \frac{1}{N^2} 
\sum_{v \in t_i} \sum_{u \in t_j} cosine(e_v,e_u)  
\end{equation}
where 
$e_v$, and $e_u$ denote the image embeddings associated with images $v$ and $u$ respectively. 

Next, to estimate the overall IEPS score for a topic model with $k$ topics, we  compute the pairwise similarities between all pairs of topics and then aggregate them into a single score. Let $S$ denote the set of all pairs of topics in the model, i.e., $S = {(t_i, t_j) \mid 1 \leq i < j \leq k}$. Then, we can define the overall IEPS score as:
\begin{equation}
\text{IEPS} = \frac{1}{|S|}\sum_{(t_i,t_j) \in S} \text{IEPS}_{\text{pair}}(t_i, t_j)
\end{equation}
This similarity metric ranges between [0,1] with 
a lower score indicating more diverse topics.

\section{Datasets}


We propose a new benchmark for  multimodal topic modeling comprising six diverse datasets. 
For illustration, Table \ref{tab:datasets_example} presents a sample document from each dataset, while the key properties of each dataset are shown in Table \ref{tab:datasets_stats}. \new{We want to alert our readers that some datasets include hateful textual and graphic elements. 
The datasets are:} 






\descr{MS COCO (Microsoft Common Objects in Context)}~\cite{lin2014microsoft} is a popular dataset for image captioning, object detection, and image segmentation tasks, 
consisting of over 200K images labeled with bounding boxes and category labels for more than 80 object categories, 
as well as \new{5} 
captions per image.  The images were sourced from Flickr and annotated by crowd workers. 
\new{We used a randomly selected subset of 30K multimodal documents (image-caption pairs)}. 


\descr{VIST (Visual Storytelling Dataset)}~\cite{huang-etal-2016-visual} is a dataset of multimodal stories comprising sequences of images with corresponding descriptions. 
All images come from Flickr, and the textual 
stories were crowd-sourced. 
For our experiments, we again randomly selected from VIST 30K multimodal documents, each consisting of an image along with the description of the portion of the story associated with the image.




\descr{T4SA 
}
~\cite{8265255}  is a large-scale \textbf{T}witter dataset 
designed for \textbf{S}entiment \textbf{A}nalysis. 
It contains textual and multimodal data obtained through 
the TwitterAPI and 
each tweet is automatically annotated with its sentiment polarity.
In our experiments, we used 
a random sample of 30K multimodal 
tweets balanced across sentiment classes.




\descr{MMHS150K}~\cite{Gomez_2020_WACV} is a hate speech dataset consisting of 150K Twitter multimodal documents. Each document was then labeled by crowdsourcing with the particular community that was attacked, such as racist, sexist, or homophobic.
Once again, we selected 30K random documents. 


\descr{HC-4chan (Hateful content on 4chan)} ~\cite{gonzalez2022understanding} contains posts, phrases, and images  containing 
hateful and discriminatory content. 
It consists of 21K images,  
identified as presenting Antisemitic/Islamophobic content by CLIP. 
We removed near-duplicate images and documents 
with no text, resulting in a 
dataset of about 18K multimodal 
documents.



\descr{MEWA (Multimodal English Wikipedia Articles)} ~\cite{zosa2022multilingual} 
comprises English Wikipedia Articles from the Wikipedia Comparable Corpora\footnote{
\scriptsize{linguatools.org/tools/corpora/wikipedia-comparable-corpora/}}, aligned with images from the Wikipedia-based Image Text dataset (WIT)~\cite{10.1145/3404835.3463257}. Each document 
consists of a complete English Wikipedia Article and its corresponding image. For our analysis, we used the publicly available subset of ~18.5K
documents\footnote{
\scriptsize{https://github.com/ezosa/M3L-topic-model/tree/master/data}}. 

\section{Experiment Setup} 


\descr{Textual Baselines:} We consider LDA~\cite{10.5555/944919.944937}, ZeroShotTM~\cite{bianchi-etal-2021-cross}, and CombinedTM~\cite{bianchi-etal-2021-pre} as strong textual baselines. For LDA, we use the  
OCTIS~\cite{terragni-etal-2021-octis} implementation. The parameters controlling the document-topic and word-topic distribution for LDA are estimated during training, as in prior work~\cite{bianchi-etal-2021-pre}. For the neural topic models, ZeroShotTM and CombinedTM, we utilize their publicly available implementations
\footnote{
\scriptsize{https://github.com/MilaNLProc/contextualized-topic-models\label{footnote:github_ctm}}}. In training these models, we employ the Adam optimizer and apply a 20\% dropout rate. 

\descr{Configurations:} We encode text and images with 
OpenAI's  CLIP~\cite{pmlr-v139-radford21a}, 
which captures content similarity across modalities. 
We use the text and image encoder of \texttt{clip-ViT-B-32}, which is available in the SBERT's library\footnote{\scriptsize{https://www.sbert.net/}}~\cite{reimers2019sentence}. For hyperparameter settings, we follow ~\cite{bianchi-etal-2021-cross} and \cite{zosa2022multilingual}.
We train models for 100 epochs, computing all the metrics for 25, 50, 75, and 100 topics. Results for each metric are averaged over 5 random seeds. The data is preprocessed following~\cite{bianchi-etal-2021-cross}. We 
restrict the vocabulary size to the top 2,000 most frequent terms to speed up computation. We remove English stopwords by using the NLTK library\footnote{\scriptsize{https://www.nltk.org/}}. We also remove punctuation and digits as suggested by prior work.

For the development of the \texttt{Multimodal-ZeroShotTM} model, we adapt ZeroShotTM~~\cite{bianchi-etal-2021-cross} and rely on the original implementation. Consistent with the original setup, our inference network structure comprises one fully connected hidden layer followed by a softplus layer with 100 dimensions.

For \texttt{Multimodal-Contrast} model, we adapt M3L-Contrast~\cite{zosa2022multilingual} and base our code on the author's original implementation\footnote{\scriptsize{https://github.com/ezosa/M3L-topic-model}}. We use a batch size of 32, set the temperature $\tau$ to 0.07, and the contrastive weight $\omega$  to 100, as in the original model's configurations.


\descr{Topic descriptors:}
VAE-based neural topic models like our \texttt{Multimodal-ZeroShotTM} and \texttt{Multimodal-Contrast} obtain the representative keywords of each topic  from the topic-vocab weight matrix used for reconstructing the BoW. 
However, \texttt{Multimodal-Contrast} does not reconstruct image features (see Figure \ref{fig:M3L_original_architecture}), so a different approach is needed to obtain the most relevant images per topic.  To this end, after model training, we rely on the document-topic distributions associated with the input documents and select the images of the 
$N$ documents with the highest contribution for each topic. We use this same approach for \texttt{Multimodal-ZeroShotTM} to ensure a fair comparison. 
 All our experiments are run considering the top 10 words and  \new{top 10 }images per topic. \new{This decision is based on prior work~\cite{bianchi-etal-2021-pre,ding-etal-2018-coherence, hoyle2021automated,li-etal-2023-diversity,10.5555/1857999.1858011}, who identified that the top 10 terms typically account for about 30\% of the topic mass, providing sufficient information to determine the subject area and distinguish one topic from another~\cite{10.5555/1857999.1858011}. We apply IEC and IEPS over 10 images, maintaining consistency with the most respected prior work.}
 
 \descr{Dataset size:}  \new{To ensure computational feasibility for our extensive experiments, 
we restricted the datasets size to 30K documents. 
Notice that testing on 30K documents considerably outnumbered assessments of previous topic modeling algorithms, such as CombinedTM~\cite{bianchi-etal-2021-pre} and our baseline, M3L-Contrast~\cite{zosa-pivovarova-2022-multilingual}, which were initially assessed using a sample of only 20K documents.}

\descr{Topic Models Overlap:} Two topic models might exhibit similar coherence and diversity, but actually generate different topics. So, as part of our systematic comparison between \texttt{Multimodal-ZeroShotTM} and \texttt{Multimodal-Contrast}, we also explore the overlap between the topics they generate. 
Specifically, given a pair of topic models,  we construct a topic similarity matrix based on the most relevant keywords using the I-RBO diversity metric \cite{bianchi-etal-2021-pre}, 
ranging in [0-1]  with higher scores indicating more substantial topic overlap. 
Then, to align the topics between models, we apply the Hungarian method~\cite{hungarian_method}. 
In the results, we report the mean $M$ and standard deviation $SD$ of the topic overlap between models 
across multiple datasets, numbers of topics, and random seeds.

\descr{Visual learned features:} In order to reconstruct image embeddings from the input,
\texttt{Multimodal-ZeroShotTM} uses a weight topic-image features matrix. After training the model, analyzing such structure can provide valuable insights 
on the relevance of specific visual attributes to each topic as well as into the neural model's limitations. 
In our experiments, we use a CLIP Guided Diffusion model ~\cite{NEURIPS2021_49ad23d1}\footnote{\scriptsize{https://github.com/nerdyrodent/CLIP-Guided-Diffusion}} to generate an image per topic given the weight topic-image feature matrix. 

\descr{User Study}: We conduct a user study with two core objectives. Firstly, we validate our proposed metrics, IEC and IEPS, ensuring their alignment with human judgments. Secondly, we perform a qualitative analysis to discern 
variations between our proposed models. Nine computer scientists participated in the study, evaluating the coherence and diversity of keyword and image sets generated by our multimodal solutions. 

For coherence evaluation, our approach, inspired by prior studies~\cite{aletras-stevenson-2013-evaluating, hoyle2021automated}, involves presenting participants with either 10 keywords or 10 images representing a topic. Participants use a 5-point Likert scale, where higher scores indicate higher similarity, to gauge the relatedness between these topics' descriptors. For diversity evaluation, participants assess the similarity between two topics at a time, each represented by a set of 10 keywords or 10 images, using the same Likert scale, with higher values indicating greater similarity (i.e., lower diversity). We restrict the evaluation to topics generated solely from the MS COCO dataset due to time constraints, supported by our pilot study's findings. 

The inter-annotator agreement (IAA) is computed as the average Spearman correlation between each respondent's scores and the averages of scores from other respondents. For coherence, the IAA is 0.71 and 0.62 for keywords and images, respectively, indicating a strong agreement among participants. For diversity, it is 0.75 and 0.84, pointing to a very strong consensus. To assess the alignment between humans and our proposed metrics, IEC, and IEPS, we calculate the mean Spearman correlation between the automatic metrics and human ratings, following the methodology suggested by prior work~\cite{aletras-stevenson-2013-evaluating, hoyle2021automated}. 

\section{Results}
\label{sec:results}

As an example, Table \ref{fig:example_topics} presents   topics that can be retrieved  from MS COCO using our multimodal solutions. 
We report the models' performance below.

\begin{table}[!h]
\setlength{\tabcolsep}{2pt}
\centering
\resizebox{\columnwidth}{!}{%
\footnotesize
\begin{tabularx}{\columnwidth}{sb}
\toprule
\multicolumn{1}{c}{Model} & \multicolumn{1}{c}{Topics descriptors} \\ \hline
\multirow{2}{*}{M-Z} & snow, person, skis, covered, mountain, ski, snowy, slope, hill, skiing  \\
 &    \includegraphics[width=0.81\columnwidth] {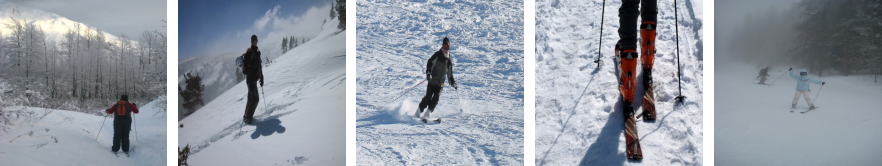}  \\ 
\midrule
\multirow{2}{*}{M-C}  & snow, covered, skis, slope, hill, snowy, ski, mountain, skiing, snowboard \\
 &  \includegraphics[width=0.81\columnwidth] {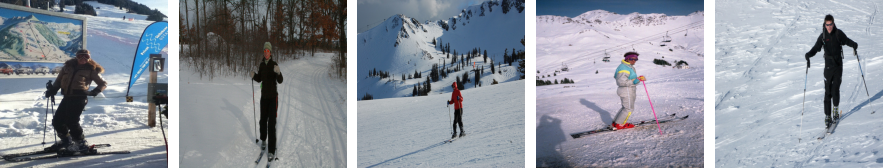}  \\
\midrule
\midrule
\multirow{2}{*}{M-Z} &mirror, toilet, bathroom, sink, wall, tub, shower, window, door, bath  \\
 &  \includegraphics[width=0.83\columnwidth] {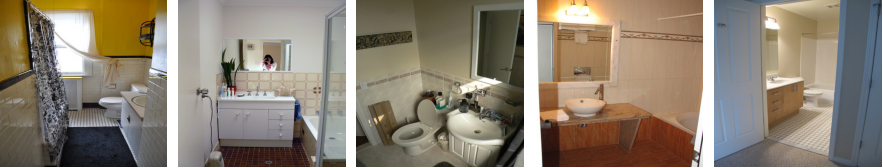}  \\
\midrule
\multirow{2}{*}{M-C} & bathroom, toilet, mirror, sink, tub, shower, bath, wall, tiled, door \\
 &  \includegraphics[width=0.83\columnwidth] {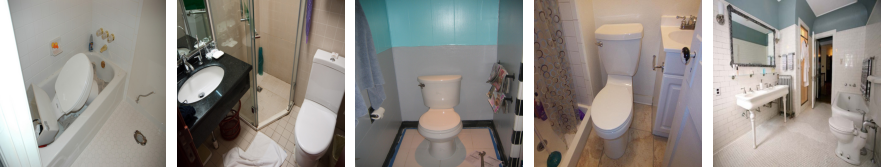}  \\
\midrule
\midrule
 \multirow{2}{*}{M-Z} &water, boat, body, beach, ocean, river, boats, lake, sandy, shore  \\
 &  \includegraphics[width=0.81\columnwidth] {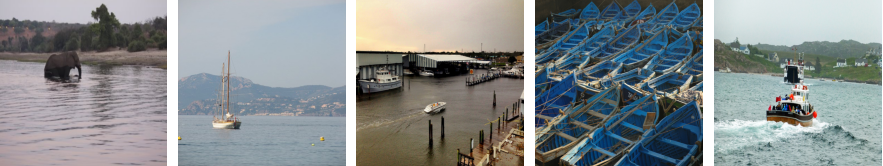}  \\
\midrule
\multirow{2}{*}{M-C} &beach, water, boat, body, ocean, boats, kite, river, kites, sandy  \\
 &  \includegraphics[width=0.81\columnwidth] {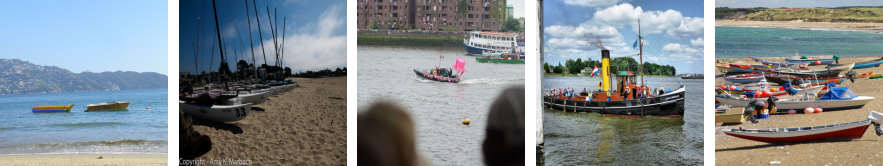}  \\
 
 \bottomrule
\end{tabularx}
}
\caption{\new{Topics retrieved using \texttt{Multimodal-ZeroShotTM} (M-Z) and \texttt{Multimodal-Contrast}  (M-C) on 
MS COCO 
} 
}
\label{fig:example_topics}
\end{table}






\descr{Topic Coherence and Diversity:}
Table \ref{tab:combined_results} 
shows the overall performance of the models in terms of  coherence and diversity of the topics' descriptors. 
\begin{table}[!htbp]
\setlength{\tabcolsep}{2pt}
\centering
\footnotesize
\resizebox{\columnwidth}{!}{
\begin{tabular}{lrrrrrrr}
\toprule
\multicolumn{1}{c}{} &
 \multicolumn{4}{c}{Coherence} & \multicolumn{3}{c}{\hspace{8pt} Diversity}  \\
  \midrule

\multicolumn{1}{c}{Metrics} &
  \multicolumn{1}{|c}{NPMI} &
  \multicolumn{1}{c}{$C_{v}$} &
  \multicolumn{1}{c}{WE} &
  \multicolumn{1}{c}{IEC} &
   \multicolumn{1}{c}{\hspace{8pt}TD} &
  \multicolumn{1}{c}{I-RBO} &
  \multicolumn{1}{c}{IEPS}  \\ 
\midrule
\multicolumn{1}{l|}{LDA} & -0.14 & .39 & .15 & & \textbf{.84} & .97 & \\
\multicolumn{1}{l|}{CombinedTM} & \textbf{.04} & \textbf{.52} & .21 & & .50 & .96 & \\
\multicolumn{1}{l|}{ZeroShotTM}  & .03 & .51 & \textbf{.22} & & .60 & \textbf{.98} & \\
\midrule
\multicolumn{1}{l|}{Multimodal-Contrast}& \textbf{.04} & .50 & \textbf{.22} & \textbf{.67} & .47 & .94 & \textbf{.41} \\
\multicolumn{1}{l|}{Multimodal-ZeroShotTM} & .03 & .51 & \textbf{.22} & .56 & .60 & \textbf{.98} &.44 \\
\bottomrule
\end{tabular}
}
\caption{Average topic coherence and diversity scores across all datasets. Top scores are bold.}
\label{tab:combined_results}
\end{table}

\begin{table*}[!h]
\setlength{\tabcolsep}{2pt}
\centering
\resizebox{\textwidth}{!}{%
\footnotesize
\begin{tabular}{lrrrr|rrrr|rrrr|rrrr|rrrr|rrrr}
\toprule
\multicolumn{1}{c}{Dataset} &
  \multicolumn{4}{c}{MS COCO} &
  \multicolumn{4}{c}{VIST} &
  \multicolumn{4}{c}{T4SA} &
  \multicolumn{4}{c}{MMHS150K} &
  \multicolumn{4}{c}{HC-4chan} &
  \multicolumn{4}{c}{MEWA} \\
  \midrule
\multicolumn{1}{c}{Metrics} &
  \multicolumn{1}{|c}{$\tau$} &
  \multicolumn{1}{c}{$\phi$} &
  \multicolumn{1}{c}{$\alpha$} &
  \multicolumn{1}{c}{$\gamma$} &
   \multicolumn{1}{|c}{$\tau$} &
  \multicolumn{1}{c}{$\phi$} &
  \multicolumn{1}{c}{$\alpha$} &
  \multicolumn{1}{c}{$\gamma$} &
   \multicolumn{1}{|c}{$\tau$} &
  \multicolumn{1}{c}{$\phi$} &
  \multicolumn{1}{c}{$\alpha$} &
  \multicolumn{1}{c}{$\gamma$} &
   \multicolumn{1}{|c}{$\tau$} &
  \multicolumn{1}{c}{$\phi$} &
  \multicolumn{1}{c}{$\alpha$} &
  \multicolumn{1}{c}{$\gamma$} &
   \multicolumn{1}{|c}{$\tau$} &
  \multicolumn{1}{c}{$\phi$} &
  \multicolumn{1}{c}{$\alpha$} &
  \multicolumn{1}{c}{$\gamma$} &
  \multicolumn{1}{|c}{$\tau$} &
  \multicolumn{1}{c}{$\phi$} &
  \multicolumn{1}{c}{$\alpha$} &
  \multicolumn{1}{c}{$\gamma$}  \\ 
  \midrule
  \multicolumn{1}{l|}{LDA} & -.10 & .37 & .14 &  & -.23 & .34 & .13 &  & -.25 & .36 & .13 &  & -.17 & .31 & .16 &  & -.21 & .40 & .13 &  & .10 & .57 & .20 &  \\
\multicolumn{1}{l|}{CombinedTM}& \textbf{.13} & .55 & .22 &  & \textbf{-.01} & \textbf{.42} & .21 &  & \textbf{.02} & \textbf{.45} & .22 &  & \textbf{.00} & \textbf{.45} & .22 &  & -.07 & \textbf{.58} & .21 &  & .13 & .64 & .20 &  \\
\multicolumn{1}{l|}{ZeroShotTM} & .12 & .54 & \textbf{.23} &  & -.02 & .39 & \textbf{.23} &  & \textbf{.02} & .44 & \textbf{.23} &  & \textbf{.00} & .44 & .22 &  & -.07 & .57 & \textbf{.22} &  & .14 & .67 & \textbf{.22} &  \\
\midrule
\multicolumn{1}{l|}{Multimodal-Contrast} & \textbf{.13} & \textbf{.56} & \textbf{.23} & \textbf{.75} & -.03 & .35 & \textbf{.23} & \textbf{.69} & -.02 & .38 & .20 & \textbf{.59} & -.01 & .40 & \textbf{.23} & \textbf{.65} & \textbf{-.02} & .54 & .19 & \textbf{.65} & \textbf{.16} & \textbf{.77} & .20 & \textbf{.68} \\
\multicolumn{1}{l|}{Multimodal-ZeroShotTM} & .12 & .54 & \textbf{.23} & .66 & \textbf{-.01} & .39 & \textbf{.23} & .56 & .01& .44 & \textbf{.23} & .47 & \textbf{.00} &\textbf{.45}  & .22 & .56 & -.07 & \textbf{.58} & .21 & .57 & .14 & .66 & .21 & .54 \\
\bottomrule
\end{tabular}
}
\caption{Topic's coherence scores per dataset. We used the following abbreviations: NPMI ($\tau$), $C_{v}$ ($\phi$), WE ($\alpha$), and IEC ($\gamma$).  Average results  over $4$ number of topics ($K = 25, 50, 75, 100$), where the results for each $K$ are averaged over $5$ random seeds. We bold the highest scores.}
\label{tab:results_coherence_per_dataset}
\end{table*}

\begin{table*}[!t]
\setlength{\tabcolsep}{2pt}
\centering
\resizebox{\textwidth}{!}{%
\footnotesize
\begin{tabular}{lrrr|rrr|rrr|rrr|rrr|rrr}
\toprule
\multicolumn{1}{c}{Dataset} &
  \multicolumn{3}{c}{MS COCO} &
  \multicolumn{3}{c}{VIST} &
  \multicolumn{3}{c}{T4SA} &
  \multicolumn{3}{c}{MMHS150KK} &
  \multicolumn{3}{c}{HC-4chan} &
  \multicolumn{3}{c}{MEWA} \\
  \midrule
\multicolumn{1}{c}{Metrics} &
  \multicolumn{1}{|c}{TD} &
  \multicolumn{1}{c}{I-RBO} &
  \multicolumn{1}{c}{IEPS} &
  \multicolumn{1}{|c}{TD} &
  \multicolumn{1}{c}{I-RBO} &
  \multicolumn{1}{c}{IEPS} &
  \multicolumn{1}{|c}{TD} &
  \multicolumn{1}{c}{I-RBO} &
  \multicolumn{1}{c}{IEPS} &
  \multicolumn{1}{|c}{TD} &
  \multicolumn{1}{c}{I-RBO} &
  \multicolumn{1}{c}{IEPS} &
  \multicolumn{1}{|c}{TD} &
  \multicolumn{1}{c}{I-RBO} &
  \multicolumn{1}{c}{IEPS} &
  \multicolumn{1}{|c}{TD} &
  \multicolumn{1}{c}{I-RBO} &
  \multicolumn{1}{c}{IEPS} 
  \\ 
  \midrule
\multicolumn{1}{l|}{LDA} & \textbf{.75} & .98 &  & \textbf{.93} & \textbf{1.00} &  & \textbf{.96} & \textbf{1.00} &  & \textbf{.85} & .90 &  & \textbf{.89} & \textbf{1.00} &  & .65 & .96 &  \\
\multicolumn{1}{l|}{CombinedTM} & .57 & .98 &  & .40 & .94 &  & .43 & .96 &  & .41 & .95 &  & .40 & .94 &  & .76 & \textbf{.99} &  \\
\multicolumn{1}{l|}{ZeroShotTM} & .67 & \textbf{.99} &  & .60 & .99 &  & .56 & .98 &  & .50 & \textbf{.97} &  & .50 & .96 &  & \textbf{.77} & \textbf{.99} &  \\ \hline
\multicolumn{1}{l|}{Multimodal-Contrast} & .65 & \textbf{.99} & \textbf{.46} & .56 & .98 & \textbf{.45} & .52 & .97 & \textbf{.28} & .29 & .87 & \textbf{.41} & .23 & .87 & \textbf{.50} & .58 & .98 & \textbf{.37} \\
\multicolumn{1}{l|}{Multimodal-ZeroShotTM} & .68 & \textbf{.99} & .47 & .60 & .99 & .48 & .56 & .98 & .34 & .50 & \textbf{.97} & .45 & .50 & .96 & .53 & \textbf{.77} & \textbf{.99} & .38 \\
%
  \bottomrule
\end{tabular}
}
\caption{Diversity scores of the top keywords and images.   Average results over $4$ number of topics ($K = 25, 50, 75, 100$), 
with results for each $K$ 
averaged over $5$ random seeds. We bold best scores.  }
\label{tab:results_diversity_per_dataset}
\end{table*}

For the metrics assessing the coherence of the textual descriptors, 
as expected, LDA performs worse, while both \texttt{Multimodal-ZeroShotTM} and \texttt{Multimodal-Contrast} perform similarly to ZeroShotTM and CombinedTM, indicating that  processing the images of the corpus does  not influence the coherence of the textual descriptors.
%
%
Interestingly, we also observe subtle differences in the NPMI and $C_{v}$ scores between \texttt{Multimodal-ZeroShotTM} and \texttt{Multimodal-Contrast}. 
As shown in Table \ref{tab:results_coherence_per_dataset}, depending on the metrics and dataset combinations, one model outperforms the other without a clear winner. For instance, \texttt{Multimodal-ZeroShotTM} can generate more coherent topics than \texttt{Multimodal-Contrast} in the VIST,  T4SA, and MMHS150 datasets but 
not 
in MS COCO and MEWA. 
As for image descriptors, 
the top relevant images of each topic identified by \texttt{Multimodal-Contrast} 
seem
more related (i.e., higher IEC) than those selected by \texttt{Multimodal-ZeroShotTM} overall and across datasets.

\begin{figure*}[h!]
\setkeys{Gin}{width=0.15\linewidth}
\includegraphics{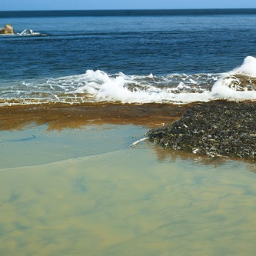}
\hfill
\includegraphics{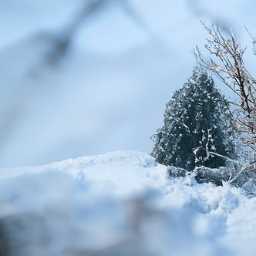}
\hfill
\includegraphics{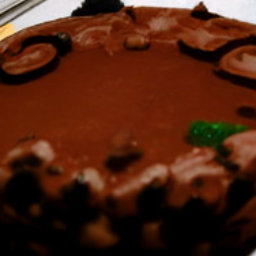}
\hfill
\includegraphics{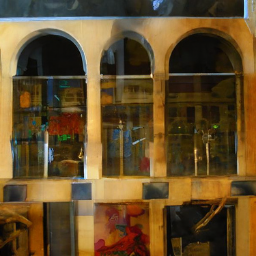}
\hfill
\includegraphics{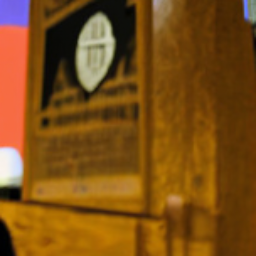}
\hfill
\includegraphics{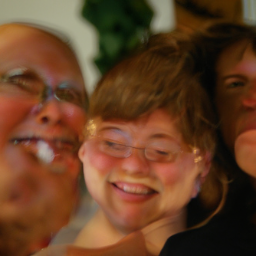}

\caption{Images 
from a CLIP-Guided Diffusion Model over the latent space. 
Topics from left to right are \{water, lake, sand, beach\}, \{snow, tree, Christmas, white\}, \{cake, made, candles, birthday\}, \{art, building, glass, amazing, architecture\}, \{students, graduation, speech, school\}, and \{family, together, happy, whole\}.}
\label{fig:learned_visual_features}
\end{figure*}
Table  \ref{tab:combined_results} and \ref{tab:results_diversity_per_dataset} report the performance of the models in terms of their ability to generate diverse topics. LDA is one of the top performers, but it scored the lowest for coherence by a wide margin (see Table  \ref{tab:combined_results} and \ref{tab:results_coherence_per_dataset}).

We also observe that for all the datasets, the keywords generated by \texttt{Multimodal-ZeroShotTM} are significantly more diverse than those generated by \texttt{Multimodal-Contrast}. 
Finally,  we find that the images representing the topics are 
more diverse (i.e., lower IEPS) in \texttt{Multimodal-Contrast} than in \texttt{Multimodal-ZeroShotTM}. 


%
Plausibly, the observed superiority of \texttt{Multimodal-Contrast} over  \texttt{Multimodal-ZeroShotTM}  in terms of image coherence and diversity can be attributed to 
their different training objectives. \texttt{Multimodal-Contrast} incorporates a Contrastive Learning loss~\cite{1467314} to maximize the similarity between positive pairs (text-image pairs that belong together) while minimizing the similarity between negative pairs. Training the model to explicitly differentiate between related and unrelated pairs may stimulate the model to learn more discriminative image representations~\cite{Yu_2022_CVPR}, which can better support similar judgments involved in creating more coherent and diverse topic models.
More speculatively, the reason why this works for image but not for text descriptors may be due to the fact that although pre-trained multimodal representations (e.g., CLIP) map 
data from different modalities into the same 
space,  embeddings from different modalities are located in separate regions~\cite{NEURIPS2022_702f4db7} and therefore could be influenced differently by Contrastive Learning.

\descr{Topic Overlap And 
Visual Features: }
Table \ref{tab:topics_overlap} shows the topic's overlap  between ZeroShotTM and our proposed 
models. 
\begin{table}[!h]
\centering
\resizebox{\columnwidth}{!}{%
\footnotesize
\begin{tabular}{llrr}
\toprule
\multicolumn{2}{c}{Models} &
  \multicolumn{1}{|c}{$M$} &
  \multicolumn{1}{c}{$SD$} \\ 
  \midrule
   
\multicolumn{1}{l|}{ZeroShotTM}&\multicolumn{1}{l|}{Multimodal-Contrast}& .22 & .16  \\
\multicolumn{1}{l|}{ZeroShotTM}&\multicolumn{1}{l|}{Multimodal-ZeroShotTM}& .50 & .23  \\
\multicolumn{1}{l|}{Multimodal-Contrast}&\multicolumn{1}{l|}{Multimodal-ZeroShotTM}& .21 & .16  \\

\bottomrule
\end{tabular}
}
\caption{Topic's overlap between models across all datasets and number of topics.}
\label{tab:topics_overlap}
\end{table}

Remarkably, while ZeroShotTM and \texttt{Multimodal-ZeroShotTM} exhibit similar performance (i.e., 
for coherence and diversity), the generated topics only partially overlap ($M$ = .50; $SD$ = .23). In other words, these two models generate some topics that are similar to each other, but also 
unique ones. 
Moreover, 
\texttt{Multimodal-Contrast} produces topics with significantly less keyword overlap when compared to both ZeroShotTM ($M$ = .22; $SD$ = .16) and \texttt{Multimodal-ZeroShotTM} ($M$ = .21; $SD$ = .16). 

Figure \ref{fig:learned_visual_features}  showcases images generated by the CLIP-Guided diffusion model considering the topic-image feature matrix $\gamma$ from \texttt{Multimodal-ZeroShotTM}, with topics extracted from 
VIST. 
%
%
%
Promisingly, the generated images seem to align well with the topic's descriptors,  
capturing 
abstract and complex concepts (like happy family).   

\descr{User Study:} Table \ref{tab:human_eval} displays human ratings for topics from our models, while Table \ref{tab:coherence_diversity_scores_annotated_data} presents corresponding automatic scores.

\begin{table}[!ht]
\setlength{\tabcolsep}{2pt}
\centering
\resizebox{\columnwidth}{!}{%
\footnotesize
\begin{tabular}{lrrrrrrrr}
\toprule
\multicolumn{1}{c}{} &
 \multicolumn{4}{c}{Coherence} & \multicolumn{4}{c}{Diversity}  \\
  \midrule
  \multicolumn{1}{c}{} & \multicolumn{2}{c}{Keywords} & \multicolumn{2}{c}{Images}  & \multicolumn{2}{c}{Keywords} & \multicolumn{2}{c}{Images}  \\
  \midrule
\multicolumn{1}{c}{Ratings scores} &
  \multicolumn{1}{c}{$M$} &
  \multicolumn{1}{c}{$SD$} &
  \multicolumn{1}{c}{$M$} &
  \multicolumn{1}{c}{$SD$} &
    \multicolumn{1}{c}{$M$} &
  \multicolumn{1}{c}{$SD$} &
    \multicolumn{1}{c}{$M$} &
  \multicolumn{1}{c}{$SD$} \\
  \midrule

Multimodal-ZeroShotTM   & \multicolumn{1}{l}{3.84}          & 1.05                   & \textbf{4.66}         & 0.60                   & \multicolumn{1}{l}{1.28}          & 0.64                   & \textbf{1.42}         & 0.95      \\
Multimodal-Contrast & \multicolumn{1}{l}{\textbf{4.05}} & 0.95                   & 4.51                  & 0.79                   & \multicolumn{1}{l}{\textbf{1.23}} & 0.67                   & 1.43                  & 0.95                   \\

  \bottomrule
\end{tabular}
}
\caption{\new{Mean and standard deviation of rating scores for evaluating coherence (higher values indicate higher coherence) and diversity (lower values indicate higher diversity) of sets of keywords and images generated by our models.}}
\label{tab:human_eval}
\end{table}

\begin{table}[!ht]
\setlength{\tabcolsep}{2pt}
\centering
\resizebox{\columnwidth}{!}{%
\footnotesize
\begin{tabular}{lrrrrrrr}
\toprule

\multicolumn{1}{c}{Metrics} &
  \multicolumn{1}{|c}{NPMI} &
  \multicolumn{1}{c}{$C_{v}$} &
  \multicolumn{1}{c}{WE} &
  \multicolumn{1}{c}{IEC}  &
  \multicolumn{1}{c}{TD} &
  \multicolumn{1}{c}{I-RBO} &
  \multicolumn{1}{c}{IEPS}  \\ 
  \midrule

\multicolumn{1}{l|}{Multimodal-ZeroShotTM} & \textbf{.09} & \textbf{.48}& .23 & \textbf{.75} & 1.00 & 1.00 &\textbf{0.43}\\
\multicolumn{1}{l|}{Multimodal-Contrast} &  .06& .44 & \textbf{.25} & .70 & 1.00 & 1.00 &0.44\\

  \bottomrule
\end{tabular}
}
\caption{ \new{Automatic coherence and diversity scores from the topics used in our 
user study. 
} }
\label{tab:coherence_diversity_scores_annotated_data}
\end{table}

To validate our metrics, IEC and IEPS, we calculated their Spearman correlation with human ratings. The results showed robust and statistically significant positive correlations: IEC (r(27) = .45, $p<.001$) and IEPS (r(27) = .44, $p<.001$), confirming the reliability of our metrics.
According to annotators, the images representing the topics in \texttt{Multimodal-ZeroShotTM} were more coherent and diverse compared to those generated by \texttt{Multimodal-Contrast}. This observation aligns with the IEC and IEPS scores for these topics (see Table \ref{tab:coherence_diversity_scores_annotated_data}), where \texttt{Multimodal-ZeroShotTM} emerged as the superior model. This result 
reinforces the credibility of our proposed metrics 
and underscores their potential for evaluating multimodal topic models. 
Human evaluators reported that \texttt{Multimodal-Contrast} generated topics with more coherent and diverse keywords compared to its multimodal counterpart, a trend supported by the WE metric. In contrast, NPMI and $C_{v}$ metrics, relying on the reference corpus (in this case, MS COCO), favored \texttt{Multimodal-ZeroShotTM}. Finally, automatic metrics indicated a tie and perfect score for topic keyword diversity, possibly due to the limited topic subsets in the user study's diversity tasks and the criteria used by TD and I-RBO,  which assess diversity based on exact keyword overlap.

\section{Conclusions and Future Work}
\label{sec:conclusions}
We present the first systematic 
evaluation of neural multimodal topic modeling, 
considering multiple non-homogeneous datasets  
and a comprehensive set of evaluation metrics. In particular, we contribute a repository of  corpora that vary in document size, source, and underlying task/domain, along with two novel metrics to assess topic image descriptors' coherence and diversity, which we validated in a preliminary user study. 
We apply the resulting evaluation framework to compare two novel multimodal topic modeling methods that we developed by adapting current SOTA architectures. Overall, our results indicate that ensemble and hybrid solutions should be explored in the future, for instance, by either merging  the output of different models or by combining different components in more complex loss functions. Leveraging GPT-like systems \cite{openai2023gpt4} is also a potential direction for future work, but first, the formidable limitation in their input size \cite{bubeck2023sparks} must be addressed.
In another short-term direction we plan to assess  whether the topic-image feature matrix $\gamma$ 
(Eq.1) can benefit 
multimodal text classification and document similarity.
\vspace{0pt}
\section*{Limitations}
As in any study, ours has limitations that need to be considered. First, we used datasets that are only available in English, which might restrict the generalizability of our findings. Moving forward, we aim to tackle this limitation by including datasets in various languages and exploring multilingual models that handle multiple languages simultaneously. 
Secondly,  we focus our topics' evaluation on their coherence and diversity. Future work should identify the quality of the results based on other aspects, such as document coverage (i.e., how well documents match their assigned topics) and topic model comprehensiveness (i.e., how thoroughly the model covers the topics appearing in the corpus). These aspects are challenging to assess when ground truth is unavailable. 
Finally, future work should explore how hyperparameters (e.g., dropout rate, weight for KL divergence loss) impact neural multimodal topic models.

\section*{Ethics Statement}


Our neural multimodal topic models are intended solely for research purposes. Any use of these models or their derived artifacts outside research contexts should be authorized accordingly. We use datasets that are publicly available. 
Users must be aware of the potential risks associated with using topic modeling algorithms. Topic models may amplify biases present in the data (e.g., if the dataset contains hateful content, the generated topics can perpetuate those discriminatory practices). Users also need to consider the biases and limitations of text and image encoders (e.g., CLIP). Moreover, neural topic models lack transparency and interpretability, meaning it becomes challenging to understand how the model arrives at particular topics.

\section*{Acknowledgements}
We would like to express our profound gratitude, first and foremost, to Aditya Chinchure, Sahithya Ravi and Raymond Li for their invaluable feedback and constructive critiques at various stages of this research. We are also thankful to Debora Nozza and Federico Bianchi for providing the initial ideas and engaging in the preliminary discussions that significantly shaped the direction of our study. Additionally, our sincere thanks go to the anonymous reviewers whose meticulous comments and suggestions have greatly enhanced the quality of this paper.

We acknowledge the support of the Natural Sciences and Engineering Research Council of Canada (NSERC). Nous remercions le Conseil de recherches en sciences naturelles et en génie du Canada (CRSNG) de son soutien.

This research was enabled in part by support provided by Calcul Québec (\url{www.calculquebec.ca}) and the Digital Research Alliance of Canada (\url{https://alliancecan.ca}).
\section*{References}

\bibliography{anthology,custom}
\bibliographystyle{lrec-coling2024-natbib}

\clearpage
\appendix

\fg{Appendices or supplementary material (software and data) will be allowed ONLY in the final, camera-ready version, but not during submission, as papers should be reviewed without the need to refer to any supplementary
materials.}
\fg{I will remove the following pages before submission. Please check if we need to move something from here to the main body of the paper}

\section{Performance of \texttt{Multimodal-ZeroShotTM} across different $\lambda$ values}

Table \ref{tab:ablation_lambda} presents the performance of \texttt{Multimodal-ZeroShotTM} for varying values of $\lambda$. This parameter adjusts the balance between the textual and image feature reconstruction losses, as defined in Equation \ref{eq:multimodalctm}. For our main experiments, $\lambda$ was set to 1. 

Our results indicate that higher  $\lambda$ values improve the coherence and diversity of the images representing a topic. However, this improvement is accompanied by a slight decrease in the coherence and diversity of the topics' most relevant keywords.

\begin{table}[ht]
\setlength{\tabcolsep}{2pt}
\centering
\footnotesize
\begin{tabular}{lrrrr|rrr}
\toprule
\multicolumn{1}{c}{} &
  \multicolumn{4}{|c}{Coherence} &
  \multicolumn{3}{|c}{Diversity} \\
  \midrule
\multicolumn{1}{c}{ } &
  \multicolumn{1}{|c}{NPMI} &
  \multicolumn{1}{c}{$C_{v}$} &
  \multicolumn{1}{c}{WE} &
  \multicolumn{1}{c}{IEC} &
  \multicolumn{1}{|c}{TD} &
  \multicolumn{1}{c}{I-RBO} &
  \multicolumn{1}{c}{IEPS} 
 \\ 
  \midrule

\multicolumn{1}{l|}{$\lambda = 1$} & \textbf{.03} & \textbf{.51} & \textbf{.22}& .56 & \textbf{.60} & \textbf{.98} & .44  \\
\multicolumn{1}{l|}{$\lambda = 60$}& \textbf{.03} & .50 & \textbf{.22} & .66 & .57 & .96 & .41   \\
\multicolumn{1}{l|}{$\lambda = 120$} & .02 & .49 & .21 & .69 & .54 & .95 & \textbf{.40} \\
\multicolumn{1}{l|}{$\lambda = 240$} & .02 & .49 & .21 & \textbf{.70} & .52 & .95 & \textbf{.40}  \\
  \bottomrule
\end{tabular}
\caption{ Quality of topics for different $\lambda$ values in \texttt{Multimodal-ZeroShotTM}. Averages are calculated across multiple datasets for $4$ number of topics ($K = 25, 50, 75, 100$), where the results for each $K$ are averaged over $5$ random seeds. Top scores are bold.}

\label{tab:ablation_lambda}
\vspace{-.5em}
\end{table}

\section{Using a Different Contextualized Representation}

We also compare the performance of neural topic models using a different text encoder. We employ the sentence-transformer model \texttt{all-mpnet-base-v2}, available in the SBERT library. Table \ref{tab:sbert_results} displays the resulting performance of the neural topic models.


\begin{table}[ht]
\setlength{\tabcolsep}{1pt}
\centering
\resizebox{\columnwidth}{!}{%
\footnotesize
\begin{tabular}{lrrrr|rrr}
\toprule
\multicolumn{1}{c}{} &
  \multicolumn{4}{|c}{Coherence} &
  \multicolumn{3}{|c}{Diversity} \\
  \midrule
\multicolumn{1}{c}{ } &
  \multicolumn{1}{|c}{NPMI} &
  \multicolumn{1}{c}{$C_{v}$} &
  \multicolumn{1}{c}{WE} &
  \multicolumn{1}{c}{IEC} &
  \multicolumn{1}{|c}{TD} &
  \multicolumn{1}{c}{I-RBO} &
  \multicolumn{1}{c}{IEPS} 
 \\ 
  \midrule

\multicolumn{1}{l|}{CombinedTM}& .03 & .51 & .21 &  & .56 & .96 &  \\
\multicolumn{1}{l|}{ZeroShotTM}& \textbf{.04} & \textbf{.52} & \textbf{.23} &  & \textbf{.58} & \textbf{.98} &  \\
\multicolumn{1}{l|}{Multimodal-Contrast}& \textbf{.04}& .51 & .22 & \textbf{.66} & .48 & .94 & \textbf{.41} \\
\multicolumn{1}{l|}{Multimodal-ZeroShotTM} &  03& .51 & .22 &.56  & \textbf{.58} & \textbf{.98}& .44  \\

  \bottomrule
\end{tabular}
}
\caption{ Performance of neural topic models using a different contextualized text encoder (i.e., \texttt{all-mpnet-base-v2}). Averages are calculated across datasets for $4$ number of topics ($K = 25, 50, 75, 100$), with each $K$  averaged over $5$ random seeds. Best scores are highlighted in bold.}
\label{tab:sbert_results}
\vspace{-.5em}
\end{table}


\fg{We won't include this on our current COLING submission. }

\section{Computing Infrastructure}
Our experiments were conducted on an NVIDIA A100 GPU, with 20 GB of memory and 12 cores of an AMD Milan 7413 processor. Although previous studies have shown that neural topic models can be run on less performant hardware, we chose this high-performance computing infrastructure to ensure efficient data processing.

\section{Runtime}
Previous research has demonstrated that vocabulary size significantly impacts the computational time of VAE-based neural topic models~\cite{bianchi-etal-2021-pre}. Consequently, in line with prior studies, we restrict the maximum number of terms in the Bag-Of-Word reconstructions to 2,000. We select the 2,000 most frequent words in the corpus for this purpose.

To compare the computational efficiency of different models, we report the time taken in seconds to complete one epoch during training. Table \ref{tab:overall_training_time} presents the required time for each neural topic model to complete one epoch. For LDA, however, we report the average training time. Our findings indicate that multimodal neural topic models require approximately 1.5 seconds more per epoch to complete compared to the unimodal ZeroShotTM model. This additional time can be attributed to the larger input sizes and the simultaneous analysis of two modalities in multimodal models. Such a difference in training time is anticipated and justifiable, considering the increased complexity of these models.

\begin{table}[!ht]
\setlength{\tabcolsep}{2pt}
\centering
\footnotesize
\begin{tabular}{lr}
\toprule
\multicolumn{1}{c}{} &
  \multicolumn{1}{c}{Training time per epoch}   \\ 
  \midrule
\multicolumn{1}{l}{LDA} & 12.49 \\
\multicolumn{1}{l}{CombinedTM} & 7.38 \\
\multicolumn{1}{l}{ZeroShotTM}& \textbf{7.27} \\
\midrule
\multicolumn{1}{l}{Multimodal-Contrast} & 8.70 \\
\multicolumn{1}{l}{Multimodal-ZeroShotTM} & 8.33 \\

  \bottomrule
\end{tabular}
\caption{Time in seconds required to complete one epoch. Averages are calculated across all datasets for 4 sets of topics ($K = 25,50,75,100$), with results for each $K$  averaged over 5 random seeds. The lowest training time is bold. }
\label{tab:overall_training_time}
\vspace{-.5em}
\end{table}


\fg{COMPLETE. Add images of examples of tasks performed}

\end{document}